\documentclass[sigconf,authorversion]{acmart}

\usepackage[labelformat=simple]{subcaption}
\usepackage{tabularx}

\usepackage{booktabs}  % professional tables

\usepackage{graphicx}
\usepackage{url}
\usepackage{balance}

\usepackage{bbm}  % for identity function \mathbb{I}
\usepackage{multirow}
\usepackage{algorithm}
\usepackage{algpseudocode}

\usepackage{hyperref}

\DeclareMathOperator*{\argmax}{arg\,max}
\DeclareMathOperator*{\argmin}{arg\,min}

\newcommand{\ie}[0]{\textit{i.e.}}
\newcommand{\eg}[0]{\textit{e.g.}}
\newcommand{\etc}[0]{\textit{etc.}}

\newcommand{\op}[1]{\textsf{#1}}

\copyrightyear{2022}
\acmYear{2022}
\setcopyright{acmcopyright}
\acmConference[CIKM '22]{Proceedings of the 31st ACM International Conference on Information and Knowledge Management}{October   17--21, 2022}{Atlanta, GA, USA}
\acmBooktitle{Proceedings of the 31st ACM International Conference on Information and Knowledge Management (CIKM '22), October   17--21, 2022, Atlanta, GA, USA}
\acmPrice{15.00}
\acmDOI{10.1145/3511808.3557441}
\acmISBN{978-1-4503-9236-5/22/10}

\newcommand{\ours}[0]{RuDi}
\newcommand{\oursfull}[0]{\textbf{Ru}le \textbf{Di}stillation}
\newcommand{\tpm}[0]{$\pm$}

\begin{document}

\title{RuDi: Explaining Behavior Sequence Models by Automatic Statistics Generation and Rule Distillation}

\author{Yao Zhang}
\email{yaozhang@fudan.edu.cn}
\author{Yun Xiong}
\authornote{Corresponding author.}
% \additionalaffiliation{\institution{Shanghai Institute for Advanced Communication and Data Science}}
\email{yunx@fudan.edu.cn}
\affiliation{%
  \institution{Shanghai Key Laboratory of Data Science, School of Computer Science, Fudan University}
  \country{China}
}

\author{Yiheng Sun}
\email{elisun@tencent.com}
\affiliation{%
  \institution{Tencent Weixin Group}
  \country{China}
}

\author{Caihua Shan}
\email{caihuashan@microsoft.com}
\affiliation{%
  \institution{Microsoft Research Asia}
  \country{China}
}

\author{Tian Lu}
\email{lutian@asu.edu}
\affiliation{%
  \institution{Department of Information Systems, W. P. Carey School of Business, Arizona State University}
  \country{United States}
}

\author{Hui Song}
\email{ivansong@tencent.com}
\affiliation{%
  \institution{Tencent Weixin Group}
  \country{China}
}

\author{Yangyong Zhu}
\email{yyzhu@fudan.edu.cn}
% \authornotemark[2]
\affiliation{%
  \institution{Shanghai Key Laboratory of Data Science, School of Computer Science, Fudan University}
  \country{China}
}

\renewcommand{\shortauthors}{Zhang et al.}

\begin{abstract}
Risk scoring systems have been widely deployed in many applications, which assign risk scores to users according to their behavior sequences. Though many deep learning methods with sophisticated designs have achieved promising results, the black-box nature hinders their applications due to fairness, explainability, and compliance consideration. Rule-based systems are considered reliable in these sensitive scenarios. However, building a rule system is labor-intensive. Experts need to find informative statistics from user behavior sequences, design rules based on statistics and assign weights to each rule. In this paper, we bridge the gap between effective but black-box models and transparent rule models. We propose a two-stage method, RuDi, that distills the knowledge of black-box teacher models into rule-based student models. We design a Monte Carlo tree search-based statistics generation method that can provide a set of informative statistics in the first stage. Then statistics are composed into logical rules with our proposed neural logical networks by mimicking the outputs of teacher models. We evaluate RuDi on three real-world public datasets and an industrial dataset to demonstrate its effectiveness.
\end{abstract}




\maketitle

\section{Introduction}
\label{sec:intro}

\begin{figure*}
\centering
\includegraphics[width=0.75\linewidth]{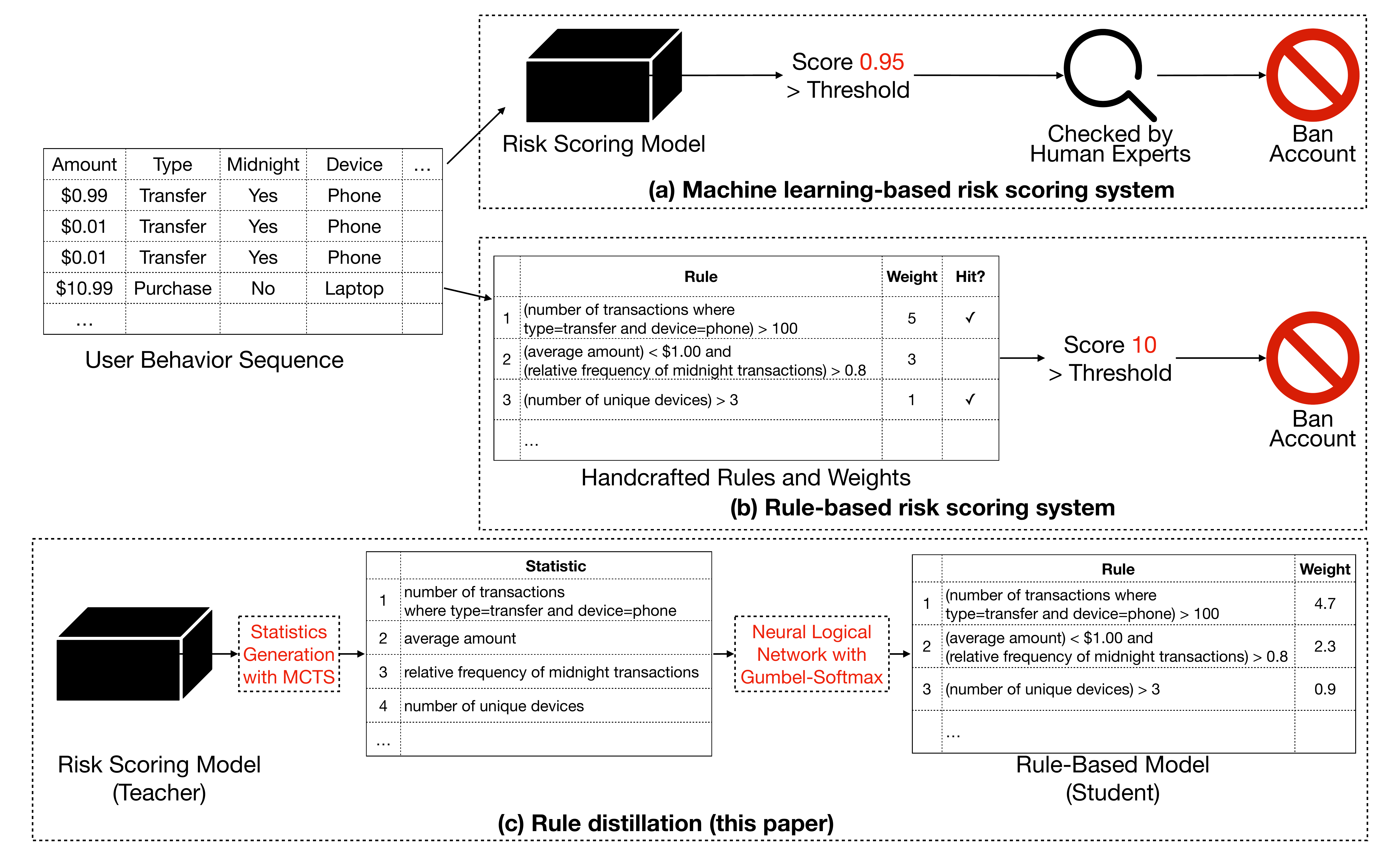}
\caption{Illustration of two types of risk scoring systems and rule distillation.
(a) A black-box risk scoring model, despite its high precision, is hardly used without human review due to fairness, explainability, and compliance requirements.
(b) Rule-based systems are considered more reliable in conservative businesses. However, designing rules are labor-intensive.
(c) In this paper, we bridge the gap by proposing a two-stage rule distillation method \ours{} that can distill knowledge of black-box models into transparent rule-based systems.
}
\label{fig:butterfly}
\end{figure*}

Exponential growth in online activities has resulted in urgent needs for anomalous accounts detection to ensure the security and reliability of cyberspace.
To evaluate risk scores of users, researchers and data scientists have proposed several approaches to model user behavior sequences, such as recurrent neural networks~\cite{yuan2019insider}, or transformers~\cite{yuan2020few}.

Though these advanced methods have achieved promising results, the nature of opacity hinders its broader application.
For example, when users of an electronic payment platform complain when their accounts have been banned, customer services may need to dig into the raw transactions and figure out if they are real fraudsters.
In conservative businesses, such as banks, the results of these black-box machine learning models are hardly used directly but needed to be reviewed by human experts first, as shown in Figure \ref{fig:butterfly}(a).
Instead, rule-based systems are considered more reliable in these scenarios, as shown in Figure \ref{fig:butterfly}(b).
A set of (weighted) rules are manually designed by human experts with their understanding and observations from historical data.
Then if a user's behavior sequence meets the rules more than a pre-defined threshold, his/her account is banned without human intervention.
Though these rules are transparent and interpretable, they are limited by experts' experiences since experts may miss some patterns.
Such systems are also hard to detect new risky patterns as fraudsters evolve.

To address the limitations of handcrafted rules, the researchers have proposed learning-based rule models \cite{wang2020transparent,wang2021scalable,angelino2017learning,yang2017scalable}.
For example, the state-of-the-art method RRL \cite{wang2021scalable} utilizes neural logical networks to learn a set of CNFs (conjunctive normal forms) or DNFs (disjunctive normal forms) from data.
But these models require inputs being fixed-length feature vectors and cannot handle sequences of user behaviors.
Moreover, due to the discrete nature of these models, they require stronger supervision signals to train. 

In this paper, we bridge the gap between effective but opaque machine learning models and transparent but simple rule systems when modeling user behavior sequences.
Instead of learning a logical rule model from raw data directly, we distill a trained risk scoring model, \ie, the teacher model, into a rule-based student model, as shown in Figure \ref{fig:butterfly}(c).
In this case, the precise predictions can be made by the teacher model, while the experts can audit the model by examining the distilled rules.
Further, the distilled rules can supplement the existing rule systems to capture new trends and provide new insights.

Despite its significance, distilling rules from a sequence model is challenging due to the following reasons:

\textbf{Exponential number of statistics.}
As Figure 1(b) illustrates, rules are typically based on some statistics that human experts handcraft.
A set of informative statistics is the cornerstone of effective rules.
However, the number of possible statistics grows exponentially as the number of conditions grows, \eg, the ``where'' clause in Figure \ref{fig:butterfly}(b).
Such a large space prevents us from enumerating all statistics, and a systematic search strategy must be applied.

\textbf{Tradeoff between effectiveness and interpretability of rule models.}
Unlike traditional knowledge distillation \cite{gou2021knowledge} where teachers and students share similar structures, rule distillation aims to use an interpretable rule model to mimic the outputs of a black-box teacher model. 
As pointed out by \cite{wang2021scalable}, traditional rule models like decision trees and rule lists/sets are hard to get comparable performance with complex models.
Tree ensembles \cite{ke2017lightgbm} though effective, are commonly not viewed as interpretable models.
On the other hand, recently proposed neural logical networks \cite{wang2020transparent,wang2021scalable} produce complicated rules where each clause may contain over dozens of literals.
These models are also hard to train due to their discrete structures.
So it is a challenge to propose an effective and interpretable rule model that can precisely mimic the outputs of teacher models.

To tackle the above challenges, we propose an effective two-stage method, named \ours{} (\oursfull{}).
In the first stage, we propose MCTS-SG, a novel \textbf{s}tatistics \textbf{g}eneration method based on \textbf{M}onte \textbf{C}arlo \textbf{T}ree \textbf{S}earch (MCTS) \cite{browne2012survey}.
We pre-define a set of basic operators like \op{Select}, \op{Mean}, \op{GroupBy}, \etc.
Then we model the statistics generation as a sequential decision process, where each statistic is composed of a series of operators.
The objective is to maximize the (expected) correlation between the teacher model's outputs and the generated statistics.
Each simulation of tree search only involves a small batch of data, and the search space of statistics is explored by MCTS systematically.
In this way, we generate a set of statistics effectively and efficiently. 

In the second stage, we learn a set of weighted rules with our proposed GS-NLN.
GS-NLN is a novel and lightweight \textbf{n}eural  \textbf{l}ogical  \textbf{n}etwork based on  \textbf{G}umbel-\textbf{s}oftmax \cite{jang2016categorical}.
Specifically, each neuron in the GS-NLN only selects two neurons from the previous layer and puts them into a conjunctive or disjunctive formula.
Unlike previous methods using logical activation functions \cite{payani2019learning} as surrogates, our proposed model directly uses boolean operations with Gumbel-softmax estimators and thus is easier to train.
Thanks to the simplicity of structures, the learned rules are more interpretable and practicable than RRL. The experimental results prove that these simple rules can achieve satisfying performance.

The main contributions of this paper are as follows:

\textbullet We propose \ours{}, a two-stage method that distills black-box behavior sequence models into rule models.
The distilled rules are interpretable and transparent, and reveal the underlying mechanisms of teacher models.
To the best of our knowledge, this is the first study on post hoc explanations of user behavior sequence models using rule distillation.

\textbullet We model the statistics generation as a sequential decision process of composing operators and design a Monte Carlo Tree Search-based search algorithm MCTS-SG.
Our approach can generate informative statistics efficiently while only requiring small batches of data during tree searching.
 
\textbullet We propose a novel and lightweight neural logical network GS-NLN.
Each neuron in the proposed model only connects to two neurons in the previous layer resulting in simpler forms of rules and better interpretability.

\textbullet Extensive experiments on three real-world datasets prove the effectiveness of our proposed \ours{}. Case studies on an industrial dataset demonstrate the interpretability of learned rules.

\section{Related Work}
\label{sec:related}
In this section, we introduce three related topics: automatic feature generation, ruled-based models and knowledge distillation.

\textbf{Automatic Feature Generation.}
Feature engineering is one of the most crucial and time-consuming steps in data mining tasks.
For tabular data, many researchers \cite{khurana2016cognito,katz2016explorekit,song2020effective,zhu2020difer} have proposed automatic feature generation methods by doing feature combination and crossing.
In \cite{kanter2015deep} the authors proposed DSM, which enumerates features in relational databases.
OneBM \cite{lam2017one} extends DSM by introducing entity graph traversal methods and selecting features with Chi-squared tests.
However, these methods cannot handle sequence data, not to mention extracting statistics from data.
They also overlook the combination explosion problems and can only generate shallow features.
There are some studies \cite{chaudhry2018motifs,chaudhry2018feature} on feature engineering with Monte Carlo tree search (MCTS) \cite{browne2012survey}, but they focus on feature selection for tabular data.

In this paper, we propose an MCTS-based statistics generation method MCTS-SG.
We use the term ``statistic'' instead of ``feature'' because we treat it as a composition of basic operators.
We systematically explore the search space of statistics with MCTS, and only small batches of data are used during searching.
By defining basic operators, our proposed MCTS-SG can generate informative statistics effectively and efficiently.

\textbf{Rule-based Models and Neural Logical Networks.}
It has been shown \cite{wang2021scalable} that traditional rule models \cite{angelino2017learning,yang2017scalable,wang2020transparent} with decision trees, rule lists, or rule sets are hard to get comparable performance.
Moreover, most methods produce deep and hierarchical IF-THEN rules that are less practical in rule-based systems.
In \cite{wang2021scalable}, the authors proposed the state-of-the-art rule-based model, RRL, which is a kind of neural network with logical activation functions \cite{payani2019learning}.
RRL is hard to train due to its discrete structures.
Besides, an extra continuous version of the network must be maintained to provide gradients. 
Though RRL learns flat rules, the number of literals in a rule is uncontrollable even with shallow network structures, leading to perplexing results.

In this paper, we propose GS-NLN, a Gumbel-softmax-based neural logical network with the constraint that each neuron can only connect to two neurons in the previous layer.
This constraint simplifies the network structure and makes the model easy to train with Gumbel-softmax estimators \cite{jang2016categorical}.
The learned rules are also neat and sound.

\textbf{Knowledge Distillation and Post-hoc Explanation with Surrogate Models.}
In \cite{hinton2015distilling} the authors proposed knowledge distillation, where a small student model is trained to mimic the outputs of a large teacher model.
Following the idea, some researchers distill the knowledge of black-box models into transparent models \cite{tan2018distill,che2015distilling}, \ie, surrogate models, as a way of post-hoc model explanation.
For example, the model auditing method \cite{tan2018distill} detects what features were used in the teacher models.
These approaches only produce importance of features.
In \cite{frosst2017distilling,liu2018improving} the authors proposed methods to distill neural networks into decision trees. 
Another branch of research [23, 24] locally fits teacher models to generate instance-wise explanations.
There is another work called rule knowledge distillation \cite{hu2016harnessing}.
It distills rules into deep neural networks, which is opposite to our setting.
Nonetheless, most existing methods cannot deal with sequence data.

In this paper, we distill a black-box sequence model into a set of rules with the help of the proposed MCTS-SG and GS-NLN.
This paper is the first study on distilling sequence models into rule systems to the best of our knowledge.

\section{Methodology}
\label{sec:method}

\subsection{Overview}

\begin{figure*}
\includegraphics[width=.7\linewidth]{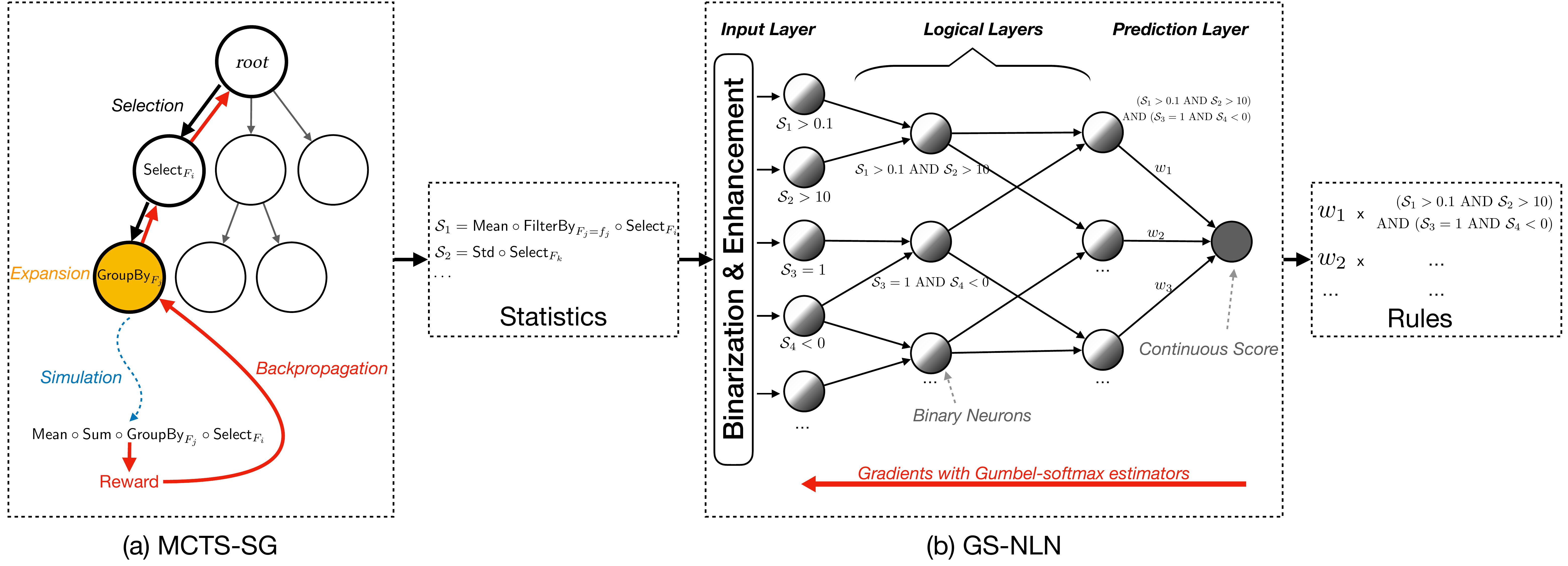}
\caption{The framework of our proposed \ours{}.
(a) Statistics generation with Monte Carlo tree search.
(b) Neural logical network with Gumbel-softmax. Here we discard disjunction parts and skip connections for simplicity.
}
\label{fig:method}
\end{figure*}

Let $\{\mathcal{X}_1, \dots, \mathcal{X}_N \}$ denote the dataset of $N$ users, where $\mathcal{X}_i$ is the sequence of user $i$'s behaviors.
All sequences contains $m_1$ categorical variables (columns) $\{F_1, \dots, F_{m_1}\}$ and $m_2$ numerical variables (columns) $\{F_{m_1+1}, \dots, F_{m_1 + m_2}\}$.
The sample spaces of categorical variables are denoted as $\mathbb{F}_j=\{f_{j1}, \dots, f_{jp}\}$, while the sample spaces of the numerical ones are $\Re$.
For notation simplicity, here we assume $|\mathbb{F}_1|=\cdots = |\mathbb{F}_{m_1}|=p$.

A trained risk scoring model, \eg, a recurrent neural network, takes the sequence $\mathcal{X}_i$ as input and produces its continuous prediction $y_i \in \Re$.
Note that we do not constrain the range of $y_i$, since we treat the risk scoring model as a black-box, and its outputs may correspond to probabilities, log-probabilities, or just real-valued risk scores.
The task of rule distillation is to find a set of weighted rules that can mimic the teacher model's outputs $y\in \Re^N$.

The framework of our proposed \ours{} is illustrated in Figure \ref{fig:method}.
\ours{} is a two-stage method.
In the first stage, we use an MCTS-based statistics generation method to generate a small set of statistics.
Then in the second stage, we binarize the statistics into literals and train a neural logical network by minimizing the ranking loss with respect to $y$.
After training, the weights and neurons in the last prediction layer are extracted as the distilled rules.

\subsection{Stage 1: Statistics Generation with MCTS}

In the first stage, we search for promising statistics that can be transformed into literals. The algorithm of MCTS-SG is summarized in Algorithm \ref{alg:mcts}.

\subsubsection{Statistics as compositions of operators}
\label{sec:ops}
In this paper, we define a statistic\footnote{We use the term statistic to represent a statistical function or the value of the function interchangeably.} as a composition of basic operators:
\begin{equation*}
    \mathcal{S} = \mathcal{T}_d \circ \cdots \circ \mathcal{T}_1,~ 2 \leq d \leq D,
\end{equation*}
where $D$ is the maximum depth of the statistics, and $\mathcal{T}_i \in \mathbb{T}$ is selected from the following pre-defined sets: 
    $\mathbb{T} = \mathbb{T}_{\text{select}}  \cup \mathbb{T}_{\text{agg}} \cup \mathbb{T}_{\text{by}} \cup \mathbb{T}_{\text{tsfm}}$.

\textbf{Select operators:}
\begin{equation*}
\mathbb{T}_{\text{select}} = \{\op{Select}_{F_j} | j=1,\dots, m_1+m_2 \}.
\end{equation*}
The operator $\op{Select}_{F_j}$ selects the $j$-th column of the data.
If $F_j$ is categorical, then values in the columns are transformed into one-hot representations, and the following operators work on each dimension of data separately.
The first operator must be from this set, \ie, $\mathcal{T}_1 \in \mathbb{T}_{\text{select}}$.

\textbf{Aggregation operators:}
\begin{equation}
\begin{aligned}
\mathbb{T}_{\text{agg}} = &\quad \{\op{Mean, Max, Min, Sum, Mean, Std, Ptp, Count, First}\}  \\
&\cup  \{\op{Percentile}_k | k=5, 10, 25, 50, 75, 90, 95 \}.
\end{aligned}
\label{eq:agg}
\end{equation}
As the name implies, an aggregation operator returns the corresponding statistics (Mean, Max, Min, Sum, Mean, Std, Ptp\footnote{Ptp: peak-to-peak}) of the input column.
\op{First} returns the first element, and \op{Count} returns the length of the sequence.
$\op{Percentile}_k$ computes the k-th percentile of the selected column.
An aggregation operator and a select operator can compose the shortest statistics.
Take $\mathcal{S}_2 = \op{Std} \circ \op{Select}_{F_k}$  in Figure \ref{fig:method}(a) as an example. $\mathcal{S}_2(\mathcal{X}_i)$ returns the standard deviation of the $k$-th column in sequence $\mathcal{X}_i$.

\textbf{By operators:} 
\begin{equation*}
\begin{aligned}
\mathbb{T}_{\text{by}} = &\quad \{\op{GroupBy}_{F_j} &&| j = 1, \dots, m_1 \} \\
& \cup \{\op{FilterBy}_{F_j=f_j}&&| j = 1, \dots, m_1, f_j \in \mathbb{F}_j  \} \\
& \cup \{\op{RetainBy}_{F_j=f_j}&&| j = 1, \dots, m_1, f_j \in \mathbb{F}_j \} \\
& \cup \{\op{SortBy}_{F_j,k} &&| j=m_1+1,\dots,m_1+m_2, k\in\{\text{asc,desc}\} \}
\end{aligned}
\end{equation*}
The $\op{GroupBy}_{F_j}$ operator partitions an input sequence into several groups according to the categorical feature $F_j$.
A $\op{GroupBy}$ operator must be followed by an aggregation operator that computes statistics within each group.
The $\op{FilterBy}_{F_j=f_j}$ ($\op{RetainBy}_{F_j=f_j}$) operator removes (keeps) rows where the categorical feature $F_j$ is $f_j$.
The $\op{SortBy}_{F_j,\cdot}$ operator sorts the input sequence in ascending/descending order.

\textbf{Transformation operators:}
\begin{equation*}
\mathbb{T}_{\text{tsfm}} = \{\op{Top}_5, \op{Abs} \}
\end{equation*}
The $\op{Top}_5$ returns the first 5 rows of the input data.
The $\op{Abs}$ returns the absolute values of the input data.

Note that our proposed method is highly extensible as users can define their own operators.
A valid statistic $\mathcal{S}(\mathcal{X}_i)$ is a scalar if a numerical column is selected, otherwise a $p$-dimensional vector.
We use $\mathbb{V}$ to denote the set of all valid statistics within the maximum depth $D$. For instance, $\op{Mean}\circ\op{Mean}\circ\op{Select}_\cdot$ is not a valid statistic.

\begin{algorithm}[t]
\caption{MCTS-SG}
\label{alg:mcts}
\begin{algorithmic}[1]
\Require Dataset $\{\mathcal{X}_i\}$, teacher model's outputs $y \in \Re^N$
\State $\hat{y} = y$
\For{$k = 1, 2, \dots, K$}
\State Initialize $\mathcal{S} \leftarrow \text{IdentityMapping}$
\For{$d = 1, 2, \dots, D$}
    \State Initialize the root node $v_0$ with $S(v_0) = \mathcal{S}$
    \For{$it = 1, 2, \dots, iterations$}
        \State Grow the tree according to Algorithm \ref{alg:mcts_sub}
    \EndFor
    \State Select the best child $v$ of the root according to Eq. \ref{eq:child}
    \State $\mathcal{T} \leftarrow A(v)$
    \state $\mathcal{S} \leftarrow \mathcal{T} \circ \mathcal{S}$
    \If{$\mathcal{S}$ is a valid statistic}
    \State break
    \EndIf
\EndFor
\State Mark $\mathcal{S}$ as invalid
\State $X_k  = \left(\mathcal{S}(\mathcal{X}_1), \cdots, \mathcal{S}(\mathcal{X}_N) \right)^\top$
\State Update $\hat{y}$ according to Eq. \ref{eq:partial}
\EndFor
\State Return $X_1, \dots, X_K$
\end{algorithmic}
\end{algorithm}

\begin{algorithm}[t]
\caption{Tree Growth}
\label{alg:mcts_sub}
\begin{algorithmic}[1]
\Require root node $v_0$, dataset $\{\mathcal{X}_i\}$, target $\hat{y}$
\State $v \leftarrow v_0$
\While{$v$ is not terminal} \Comment{Selection}
\If{$v$ is not fully expanded} \Comment{Expansion}
    \State Choose an untried valid operator $\mathcal{T}$
    \State Create a child node $v'$ of $v$ with $Q(v') = C(v') = 0, A(v')= \mathcal{T}, S(v')= \mathcal{T}\circ S(v)$ ; $v \leftarrow v'$
    \State break
\Else
    \State Select the child $v'$ according to Eq. \ref{eq:uct}; $v \leftarrow v'$
\EndIf
\EndWhile
\If{$v$ is terminal} \Comment{Simulation}
    \State $\mathcal{S} \leftarrow S(v)$
\Else
    \State Sample a valid statistic $\mathcal{S}$ according to Eq. \ref{eq:rollout}
\EndIf
\State Sample a batch index set $\mathbb{B}$
\State Compute reward $r$ according to Eq. \ref{eq:reward1+} or Eq. \ref{eq:reward2+}
\While{$v$ is not root} \Comment{Backpropagation}
    \State $C(v) \leftarrow C(v) + 1$
    \State $S(v) \leftarrow S(v) + r$
    \State $v \leftarrow parent(v)$
\EndWhile
\end{algorithmic}
\end{algorithm}

\subsubsection{MCTS Formulation}

Now we focus on generating one statistic alone.
We formulate the statistics generation as a sequential decision process.
Assume that we are going to generate a statistic $\mathcal{S}$, and we have generated a partial sequence of operators $\mathcal{T}_1, \cdots, \mathcal{T}_{i-1}$.

To determine the next optimal operator $\mathcal{T}_i$, we firstly initialize a search tree with a single root node $v_0$.
Each node $v$ (except the root) has four fields: the state $S(v)$, the action $A(v)$, the cumulative reward $Q(v)$ and the number of visits $C(v)$.
The root node only has  state $S(v_0) =  \mathcal{T}_{i-1} \circ \cdots \circ \mathcal{T}_{i}$.

Then we repeat the following steps to grow the tree as summarized in Algorithm \ref{alg:mcts_sub}.

\textbf{(Selection)} 
From the root, we select the best child iteratively until we reach a not fully expanded node or a terminal node.
A node $v$ is not fully expanded if there exists valid operator $\mathcal{T} \in \{\mathcal{T} | \exists \mathcal{T}'_{1}, \dots, \mathcal{T}'_{d} \text{ s.t. } \mathcal{T}'_{d} \circ \cdots \circ \mathcal{T}'_{1} \circ \mathcal{T} \circ S(v) \in \mathbb{V}  \}$ that is not the child of $v$.
A node is terminal if its state $S(v)$ corresponds to a valid statistic.
The best child is selected according to the UCT algorithm \cite{kocsis2006bandit}:
\begin{equation}
    \argmax_{v' \in Children(v)} \frac{Q(v')}{C(v')} + c\cdot \sqrt{\frac{2\ln C(v)}{C(v')}},
    \label{eq:uct}
\end{equation}
where $v$ is the current node (initialized as $v_0$) and $c=1/\sqrt{2}$ is the exploration constant.
The first term in Eq. \ref{eq:uct} is the estimated state value, and the second term gives high values when a child is less frequently visited.
In this case, we balance exploitation and exploration.

\textbf{(Expansion)}
If the selected node $v$ is not fully expanded, we pick an untried valid operator $\mathcal{T} \in \mathbb{T}$.
We create a new child node $v'$ of $v$ with $Q(v')=C(v')=0$, $A(v')=\mathcal{T}$, and $S(v') = \mathcal{T} \circ S(v)$.
Then we return $v = v'$.

\textbf{(Simulation)}
Now we estimate the reward of the selected node $v$.
We denote the state of the node, \ie, statistic, as $\mathcal{S} = S(v)$.
If the selected node $v$ is not a terminal node, \ie, $S(v)=\mathcal{T}_{i+j} \circ \cdots \circ \mathcal{T}_1$ is not a valid statistic,
then we randomly sample operators to form a valid one:
\begin{equation}
\mathcal{S} \leftarrow \mathcal{T}_D \circ \cdots \circ \mathcal{T}_{i+j+1} \circ \mathcal{T}_{i+j}\circ \cdots \circ \mathcal{T}_1.
\label{eq:rollout}
\end{equation}
Since we want to generate a statistic relevant to the teacher model's outputs, we design the reward function as the (absolute) correlation between them:
\begin{equation}
    r = |\text{corr}(x_{\mathbb{B}}, y_{\mathbb{B}})|,
    \label{eq:reward1}
\end{equation}
where $\mathbb{B}$ is a randomly sampled batch of indices with $|\mathbb{B}| = B$.
The statistics and target values are $x_{\mathbb{B}} = \left(\mathcal{S}(\mathcal{X}_{b(1)}), \dots, \mathcal{S}(\mathcal{X}_{b(B)})\right)^\top \in \Re^B $ and
$y_{\mathbb{B}}  = \left(y_{b(1)}, \dots, y_{b(B)}\right)^\top \in \Re^B$, where $b(i)$ is the index of the $i$-th data in the batch.
And $\text{corr}(x, y) = \frac{\bar{x}^\top \bar{y}}{\sqrt{\bar{x}^\top \bar{x} \cdot \bar{y}^\top \bar{y}}}$ is the Pearson correlation coefficient, where $\bar{x}, \bar{y}$ are centered $x,y$.

If the \op{Select} operator selects a categorical feature, the output is denoted as 
\begin{equation*}
    X_{\mathbb{B}} = \left(\mathcal{S}(\mathcal{X}_{b(1)}), \dots, \mathcal{S}(\mathcal{X}_{b(B)}) \right)^\top \in \Re^{B\times p}.
\end{equation*}
Then we define the reward as the absolute value of the coefficient of multiple correlation \cite{allison1999multiple}:
\begin{align}
& r = |\text{corr}(y', y_{\mathbb{B}})|, \label{eq:reward2}\\
\text{where} \quad& \beta, \beta_0 = \argmin_{\beta \in \Re^p, \beta_0 \in \Re} \|y_{\mathbb{B}} - X_{\mathbb{B}} \cdot \beta + \beta_0 \|_2, \nonumber \\
& y' = X_{\mathbb{B}} \cdot \beta + \beta_0. \nonumber
\end{align}
In this case we compute the Pearson correlation coefficient between the actual values and the fitted values of a linear regression model.

\textbf{(Backpropagation)}
Once we obtain the reward of the node $v$, we can back propagate the reward to its ancestors.
Specifically, do the following updates until we reach the root node:
\begin{equation*}
\begin{aligned}
    C(v) &\leftarrow C(v) + 1,\\
    Q(v) &\leftarrow Q(v) + r,\\
    v &\leftarrow parent(v).
\end{aligned}
\end{equation*}

After a certain number of iterations, we can terminate the tree growing algorithm and return the best child of the root:
\begin{equation}
v = \argmax_{v' \in Children(v_0)} \frac{Q(v')}{C(v')},
\label{eq:child}
\end{equation}
We update the partial solution with $\mathcal{T}_i = A(v)$.
Then we create a new search tree with the root $\mathcal{T}_{i} \circ \cdots \circ \mathcal{T}_{1}$ and repeat the algorithm until we construct a valid statistic.

\subsubsection{Generating top-$K$ statistics}

In the previous subsection, we describe the algorithm for generating one statistic.
Can we invoke the algorithm $K$ times to generate $K$ statistics?
Apparently no.
The algorithm's objective is to maximize the absolute correlation between the statistic and the teacher model's outputs.
If we naively run the algorithm multiple times, we are likely to obtain multiple highly correlated and redundant statistics, for example, the mean, the sum, and the 50th percentile of ``amount'' in Figure 1.

So in this subsection, we introduce two modifications to improve the diversity of the generated statistics.
First, once we obtain a statistic $\mathcal{S}_{k}$, we mark it invalid by updating $\mathbb{V} \leftarrow \mathbb{V} \backslash \{\mathcal{S}_k \}$.
In this case, it would not be generated again even in simulation processes (Eq. \ref{eq:rollout}).

Before generating the $(k+1)$-th statistic, we change the target values with:
\begin{align}
&\hat{y} = y - \left(\sum_{i=1}^{k} X_i \cdot \beta_i + \beta_0 \right), \label{eq:partial}\\
\text{where}\quad & \beta_0, \beta_1, \dots, \beta_k = \argmin_{\beta_0,\dots \beta_k } \left\|y - \left(\sum_{i=1}^{k} X_i \cdot \beta_i + \beta_0\right) \right\|_2, \nonumber\\
&X_i  = \left(\mathcal{S}_i(\mathcal{X}_1), \cdots, \mathcal{S}_i(\mathcal{X}_N) \right)^\top \in \Re^{N} \text{ or } \Re^{N \times p}, \nonumber \\
& y = \left(y_1, \dots, y_N \right). \nonumber
\end{align}
Here $X_i \in \Re \text{ or } \Re^{N \times p}$ and $\beta_i \in \Re \text{ or } \Re^p$ depend on whether the statistic acts on a categorical column.
Then we replace $y_{\mathbb{B}}$ in Eq. \ref{eq:reward1} and Eq. \ref{eq:reward2} by $\hat{y}_{\mathbb{B}}$:
\begin{align}
r &= |\text{corr}(x_{\mathbb{B}}, \hat{y}_{\mathbb{B}})|, \text{ (for numerical columns)}\label{eq:reward1+}\\
r &= |\text{corr}(\hat{y}', \hat{y}_{\mathbb{B}})|,\text{ (for categorical columns)} \label{eq:reward2+}
\end{align}

Now the objective of the algorithm is to maximize the (absolute) partial correlation between the statistic and the outputs of the teacher, given existing generated statistics.
In this case, we reduce the collinearity of statistics.

\subsubsection{Discussion}

In Eqs. \ref{eq:partial}, \ref{eq:reward1+} and \ref{eq:reward2+} we assume the linearity of data.
We can extend these formula into more general cases by considering dependency rather than linear correlation.
For example, we can replace the reward function in Eq. \ref{eq:reward1+} with the following equation
\begin{equation*}
    r = \text{CondDep}(x_{\mathbb{B}}, y_{\mathbb{B}} | X_{1,\mathbb{B}}, \dots, X_{k, \mathbb{B}}),
\end{equation*}
where $\text{CondDep}$ corresponds to some conditional dependency tests \cite{yu2020causality}, and $X_{i, \mathbb{B}}$ are values of statistics on the batch.
Moreover, the general idea of Algorithm \ref{alg:mcts} and Eq. \ref{eq:partial} is similar to the forward selection phase of forward-backward causal feature selection methods \cite{yu2020causality}.
It is also possible to incorporate the backward elimination phase to remove unsatisfying statistics generated in early steps.

\subsubsection{Complexity}

During the tree search, there are $K \times D \times B \times \text{(number\_of\_simulations)}$ data points being evaluated, which is independent of the dataset size $n$ or the space of statistics.
Since to evaluate the statistics of all data points $\mathcal{X}_i$, another $K \times N$ evaluations are inevitable,  Eq. \ref{eq:partial} only introduces little computational overhead, \ie, a regression problem with linear complexity w.r.t $N$.
We use MCTS to explore the space of statistics systematically, and Algorithm \ref{alg:mcts} is highly scalable.

\subsection{Stage 2: Rule Distillation with Neural Logical Network}

In this section, we describe how to distill rules from the teacher model with the generated statistics.
As demonstrated in Figure \ref{fig:method}(b), our proposed GS-NLN contains an input layer, several logical layers, and a prediction layer.

\subsubsection{Input Layer}
Assume that we have obtained the values of top-$K$ statistics:
\begin{equation*}
    X = (X_1, \dots, X_K) \in \Re^{N \times P},
\end{equation*}
where $P$ is the corresponding dimension after concatenation.
Some columns of $X$ have already been binary, \eg, categorical features with \op{Max} operator, while the others have not.
So we iterate over the columns of $X$.
If a column corresponds to a continuous statistic, we binarize it according to its percentiles:
\begin{equation}
x'_j = \begin{cases}
x_j \quad \text{ if }  x_j \in \{0, 1\}^N,\\
(\mathbb{I}_{x_j > x_j^{0}}, \dots, \mathbb{I}_{x_j > x_j^{90}}, \mathbb{I}_{x_j > p_j^{100}}) \in \{0, 1\}^{N \times 11} ~ \text{ otherwise.}
\end{cases}
\label{eq:binarize}
\end{equation}
Here, $\{x_j^{k}| k=0, \dots, 90, 100\}$ are percentiles of $x_j$, $\mathbb{I}_{\text{cond}}$ is the element-wise indicator function that returns 1 if $\text{cond}$ is true, otherwise 0.
We denote the binarized features as $X' = (x'_1, \dots, x'_P)\in\{0, 1\}^{N\times P'}$, and each element in $X'$ corresponds to a boolean literal.

Now let us focus on a certain row of $X'$, \ie, the boolean representations of the $i$-th user, denoted as $z_i = \text{row}_i(X')^\top \in \{0, 1\}^{P'}$.
We enhance the input by 
\begin{equation*}
    z_i^0 = (z_i^\top, (1-z_i)^\top, 1, 0)^\top \in \{0, 1\}^{h_0},
\end{equation*}
where $h_0 = 2\cdot P' + 2$.
Since the following logical layers are only cable of doing AND/OR operations, we enhance the input by appending its negations.
Moreover, extra 1 and 0 are added to enable the following layers to learn shortcuts.

\subsubsection{Logical Layers}

Our proposed GS-NLN has $L$ logical layers, and each is composed of a conjunction part and a disjunction part.
Let $z_i^{l-l} \in \{0, 1 \}^{h_{l-1}}$ denote the binary hidden representation after the $(l-1)$-th layer.
The $j$-th neuron in the conjunction part would select two elements in $z_i^{l-l}$ by sampling from the parameterized categorical distributions:
\begin{equation}
q_j^{l,k} \sim \text{onehot}\left(\text{softmax}(w^{l,k}_j)\right), ~ k=1, 2 \\
\label{eq:softmax}
\end{equation}
where $w^{l,k}_j\in \Re^{h_{l-1}}$ are parameters.
Then the value of the neuron is the boolean AND between the input values:
\begin{equation*}
    u_j^l = (q_j^{l, 1})^\top z^{l-1}_i  \cdot (q_j^{l, 2})^\top z^{l-1}_i \in \{0, 1\},
\end{equation*}

Similarly, neurons in the disjunction part conduct the boolean OR operations:
\begin{equation}
\begin{aligned}
v^l_j = 1 - (1 - (q_j^{l,3})^\top z^{l-1}_i)  \cdot (1 - (q_j^{l,4})^\top z^{l-1}_i ) \in \{0, 1\},\\
q_j^{l,k} \sim \text{onehot}\left(\text{softmax}(w^{l,k}_j)\right), ~ k=3, 4.
\end{aligned}
\label{eq:softmax2}
\end{equation}

The sampling operations in Eq. \ref{eq:softmax} and Eq. \ref{eq:softmax2} cause the network no longer differentiable.
So we resort to the Gumbel-softmax estimator \cite{jang2016categorical}, and re-write them into
\begin{equation}
\begin{aligned}
q_j^{l, k} = \text{softmax}((w^{l,k}_j + g_k)\cdot \tau^{-1}), ~ k=1,2,3,4,\\
\end{aligned}
\label{eq:gumbel}
\end{equation}
where $g_k \in \Re^{h_{l-1}}$  are $h_{l-1}$ i.i.d. samples from the Gumbel distribution $\text{Gumbel}(0, 1)$, and $\tau$ is the temperature.
As $\tau$ approaches 0, Eq. \ref{eq:gumbel} becomes identical to Eq. \ref{eq:softmax} and Eq. \ref{eq:softmax2}.
During training, we start at a high temperature $\tau=1$ and anneal to a low temperature, \eg, $\tau=0.0001$.

The output of the layer is composed of
\begin{equation*}
    z^l = (u_1, \dots, u_H , v_1, \dots, v_H, (z^{l-1})^\top)^\top \in \{0, 1\}^{h_l},
\end{equation*}
where $h_l = 2\cdot H + h_{l-1}$, and $H$ is the hidden size. We also use skip connections so that the model can adjust the complexity of rules adaptively. 
But at the last layer, we do not use the skip connection and use $\frac{R}{2}$ neurons in conjunction and disjunction parts respectively, where $R$ is the number of desired rules.

\subsubsection{Prediction Layer}
After $L$ logical layers, we obtain $z_i^L \in \{0, 1\}^{R}$ where each element represents a logical rule.
Now we get the score of the instance by a linear prediction layer $\tilde{y}_i = w^\top z_i^L$,
where $w \in \Re^R$ is the model parameter representing the weights of rules.
It is possible to pass $w$ or $\tilde{y}_i$ through non-linear functions.
For example, we could use the softplus function if we want the weights of rules nonnegative, or use sigmoid function to constrain scores to be with range (0, 1).
We leave this for future work.

\subsubsection{Model Training}
The training objective is to mimic the outputs of the teacher model.
We use the following ranking-based loss function
\begin{equation*}
    \mathcal{L} = \frac{1}{N\times(N-1) } \sum_{i\neq j} \left( \mathbb{I}_{y_i > y_j} \ln \sigma(\tilde{y}_i - \tilde{y}_j)  + \mathbb{I}_{y_i \leq y_j} \ln\sigma(\tilde{y}_j - \tilde{y}_i)  \right),
\end{equation*}
where $\sigma(\cdot)$ is the sigmoid function.
In practice, we use the stochastic gradient descent to optimize the loss function, and an instance $i$ is only compared to the samples in the same mini-batch.

\subsubsection{Differences between GS-NLN and RRL}
\label{sec:compare}

RRL \cite{wang2021scalable} is the state-of-the-art algorithm that learns logical rules.
Though our proposed GS-NLN and RRL have similar outputs, they are highly different.
RRL aims to learn rule-based representations for interpretable classification, while GS-NLN is to distill the knowledge of a teacher into a rule system.
So in GS-NLN, a neuron can only connect to two input units to maintain the low complexity of the rules.
On the contrary, in RRL a neuron can connect to any number of units for better performance.

Another difference lies in the way discrete activations are handled in networks.
In RRL, an extra continuous version of the network is maintained, where boolean operations are implemented with logical activation functions \cite{payani2019learning}.
The authors proposed the gradient grafting technique, and they use the grafted gradients of the continuous network to guide the updates of the discrete one.
In our work, since each neuron only connects to two units, we would hardly face notorious vanishing gradient problems of logical activation functions, and straightforward Gumbel-softmax estimators can learn parameters very well.

\section{Experiments}

In this section, we prove the effectiveness of our proposed \ours{}.
The codes of \ours{} and pre-processing scripts can be found at \url{https://github.com/yzhang1918/cikm2022rudi}.
Processed datasets and trained teacher models with train/valid/test splits are also included.

\subsection{Datasets and Teacher Models}

\begin{table}[h]
\caption{Statistics of datasets.}
\label{tab:dataset}
\centering
\begin{tabular}{l|cccc}
\toprule
& VEWS & Elo & RedHat & Industrial\\
\midrule
\# Users & 33,576 & 201,917 &  144,639 & 222,306\\
\# Records & 772,530 & 19,249,694  &  1,889,213 &  34,722,334\\
\# Cat. Features & 7 & 12 & 11  & 6\\
\# Num. Features & 0 & 1 & 0 & 1 \\
\bottomrule
\end{tabular}
\end{table}

\begin{table}[!t]
\centering
\caption{Default values of hyper-parameters in \ours. }
\label{tab:deafult}
\begin{tabular}{c|c}
\toprule
Hyper-parameter & Value \\
\midrule
(MCTS-SG) & \\
Statistics Depth $D$ & 4 \\
Number of statistics $K$ & 20 \\
Batch size $B$ & 128 \\
Number of simulations & 500 \\
\midrule
(GS-NLN) & \\
Number of rules $R$ & 20 \\
Hidden size $H$ & 20 \\
Number of logical layers $L$ & 2 \\
Optimizer & Adam \\
Batch size & 128 \\
Epochs & 500 \\
Learning rate & $0.1 \rightarrow 0.001$ (linear decay) \\
Temperature $\tau$ & $1.0 \rightarrow 0.0001$ (linear decay) \\
 \bottomrule
\end{tabular}
\end{table}

We use 3 public datasets:
  (1) VEWS \cite{kumar2015vews} contains 33,576 users and their edit behaviors on Wikipedia, among which Wikipedia administrators blocked 17,027 users for vandalism.
  (2) Elo\footnote{\url{https://www.kaggle.com/c/elo-merchant-category-recommendation/}} contains transactions of 201,917 users, and 97,609 users are marked positive due to their customer loyalty.
  (3) RedHat\footnote{\url{https://www.kaggle.com/c/predicting-red-hat-business-value/}} contains activities of 144,639 users, and 62,115 users are positive considering their potential business values.
An extra industrial dataset is analyzed in Section \ref{sec:case}.
The statistics of the datasets are summarized in Table \ref{tab:dataset}.

Two models are used as teachers:
(1) GRU: We use a bi-directional 2-layer GRU with 128 hidden units and dropout ratio 0.2.
(2) LightGBM: As one of the most popular methods in data mining competitions, we use LightGBM \cite{ke2017lightgbm} as the second teacher model. We use all aggregation operators introduced in Eq. \ref{eq:agg} to transform raw sequences into tabular forms.

We use 80\% users' sequences as the training data, among which 1,000 sequences are used for validation.
The remaining are used for testing.
We train GRU and LightGBM with early stopping.

Unless otherwise stated, we use the default hyper-parameters of \ours{} summarized in Table \ref{tab:deafult}.
Experiments on public datasets were conducted on a single server with 72 cores, 128GB memory, and four Nvidia Tesla V100 GPUs.

\subsection{Experiment 1: Statistics Generation}

In the first group of experiments, we test if our proposed MCTS-SG can generate informative statistics.

As we reviewed in Section \ref{sec:related}, most automatic feature engineering methods cannot be directly applied to behavioral sequences.
We consider the following alternative approaches.
(1) Random: We generate $K$ statistics by randomly composing operators introduced in Section \ref{sec:ops}.
(2) Lasso \cite{efron2004least}: We apply all aggregation operators (Eq. \ref{eq:agg}) on the datasets to transform sequences into fixed-length feature vectors, then train Lasso models to perform feature selection.
(3) LightGBM \cite{ke2017lightgbm}: Similarly, we train LightGBM on the flattened feature vectors and then select features with the highest importance scores.
(4) OneBM \cite{lam2017one}: This method can generate features on relational databases and use Chi-squared testing to select features. We transform user sequences into two relational tables to meet its input requirements.
Other feature generation methods on relational databases like DSM \cite{kanter2015deep} have been shown inferior to OneBM and thus are omitted here.

Since we use LightGBM as a baseline here, we choose GRU as the teacher model.
We measure the coefficients of multiple correlation as Eq. \ref{eq:reward2} between generated statistics/features and the teacher model's outputs on the training splits.
The higher the value is, the more informative the generated statistics are.

The results are shown in Table \ref{tab:exp_stat}. 
As we can see, the proposed MCTS-SG outperforms baselines significantly.
Lasso, LightGBM and OneBM cannot generate deep statistics as MCTS-SG and thus only achieve suboptimal performance.
Though Random can generate deep statistics, the exponentially large search space makes it impossible to generate useful statistics without systematic search strategies.
These results prove the effectiveness of our MCTS-based statistics generation method.

\begin{table}[!t]
\caption{Experimental results of statistics generation.}
\label{tab:exp_stat}
\centering
\begin{tabular}{cccc}
\toprule
& VEWS & Elo & RedHat\\
\midrule
Random   &0.7048 &0.4198 &0.5906 \\
Lasso    &0.9302 &0.6363 &0.6229 \\
LightGBM &0.9240 &0.6769 &0.6292 \\
OneBM    &0.9120 &0.6868 &0.6555 \\
\midrule
MCTS-SG (This paper) & \textbf{0.9429} &\textbf{0.7262} &\textbf{0.7872} \\
\bottomrule
\end{tabular}
\end{table}

\begin{table*}[!t]
\caption{Experimental results of rule distillation. %
}
\label{tab:exp_main}
\centering
\begin{tabular}{@{}l|cc|cc|cc@{}}
\toprule
& \multicolumn{2}{c|}{VEWS} & \multicolumn{2}{c|}{Elo} & \multicolumn{2}{c}{RedHat} \\
& Fidelity & AUC & Fidelity & AUC & Fidelity & AUC  \\
\midrule
(Teacher: GRU) & & (0.9764)  & & (0.6345)  && (0.6381) \\
Lasso         &0.8863\tpm 0.0001 &0.9637\tpm 0.0001 &0.6828\tpm 0.0003 &0.5491\tpm 0.0001 &0.7347\tpm 0.0011 &0.5828\tpm 0.0003  \\
Lasso (Binary)&0.8874\tpm 0.0001 &0.9615\tpm 0.0001 &0.7166\tpm 0.0001 &0.5763\tpm 0.0001 &0.7249\tpm 0.0023 &0.5769\tpm 0.0008  \\
CART          &0.8203\tpm 0.0001 &0.9457\tpm 0.0001 &0.7363\tpm 0.0001 &0.5784\tpm 0.0001 &0.7510\tpm 0.0001 &0.5920\tpm 0.0001  \\
RRL           &0.7901\tpm 0.0055 &0.9149\tpm 0.0069 &0.7268\tpm 0.0038 &0.5619\tpm 0.0041 &0.7621\tpm 0.0037 &0.5911\tpm 0.0045  \\
\midrule
\ours{} (This paper) &\textbf{0.9206}\tpm 0.0020 &\textbf{0.9666}\tpm 0.0006 &\textbf{0.7625}\tpm 0.0004 &\textbf{0.5807}\tpm 0.0007 &\textbf{0.8007}\tpm 0.0020 &\textbf{0.6095}\tpm 0.0013  \\
% \midrule
\midrule
(Teacher: LightGBM) & & (0.9726)  & & (0.6211)   && (0.6895)  \\
Lasso         &0.8865\tpm 0.0007 &0.9673\tpm 0.0001 &0.8619\tpm 0.0001 &0.6114\tpm 0.0001 &0.6881\tpm 0.0008 &0.5883\tpm 0.0001 \\
Lasso (Binary)&0.8721\tpm 0.0005 &0.9647\tpm 0.0001 &0.8789\tpm 0.0004 &0.6135\tpm 0.0001 &0.6870\tpm 0.0022 &0.5868\tpm 0.0011 \\
CART          &0.8409\tpm 0.0001 &0.9477\tpm 0.0001 &0.8858\tpm 0.0001 &0.6137\tpm 0.0001 &0.6977\tpm 0.0001 &0.5879\tpm 0.0001 \\
RRL           &0.8215\tpm 0.0066 &0.9234\tpm 0.0091 &0.8755\tpm 0.0071 &0.6131\tpm 0.0021 &0.7002\tpm 0.0052 &0.6030\tpm 0.0072 \\
\midrule
\ours{}  (This paper) &\textbf{0.9161}\tpm 0.0004 &\textbf{0.9680}\tpm 0.0003 &\textbf{0.9048}\tpm 0.0004 &\textbf{0.6181}\tpm 0.0006 &\textbf{0.7464}\tpm 0.0032 &\textbf{0.6198}\tpm 0.0017 \\
\bottomrule
\end{tabular}
\end{table*}

\begin{figure}
\begin{tabular}{@{}r@{}l@{}}
\begin{subfigure}{.45\linewidth}
\centering
\includegraphics[width=\linewidth]{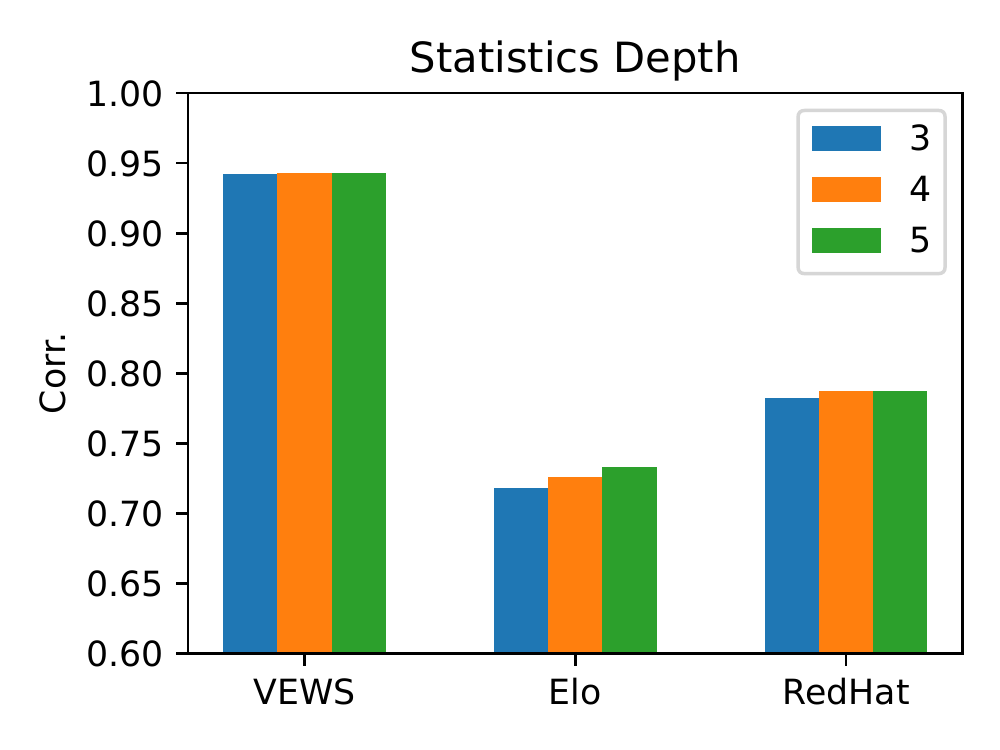}
\caption{Max statistics depth $D$}
\label{fig:exp_depth}
\end{subfigure}
&
\begin{subfigure}{.45\linewidth}
\centering
\includegraphics[width=\linewidth]{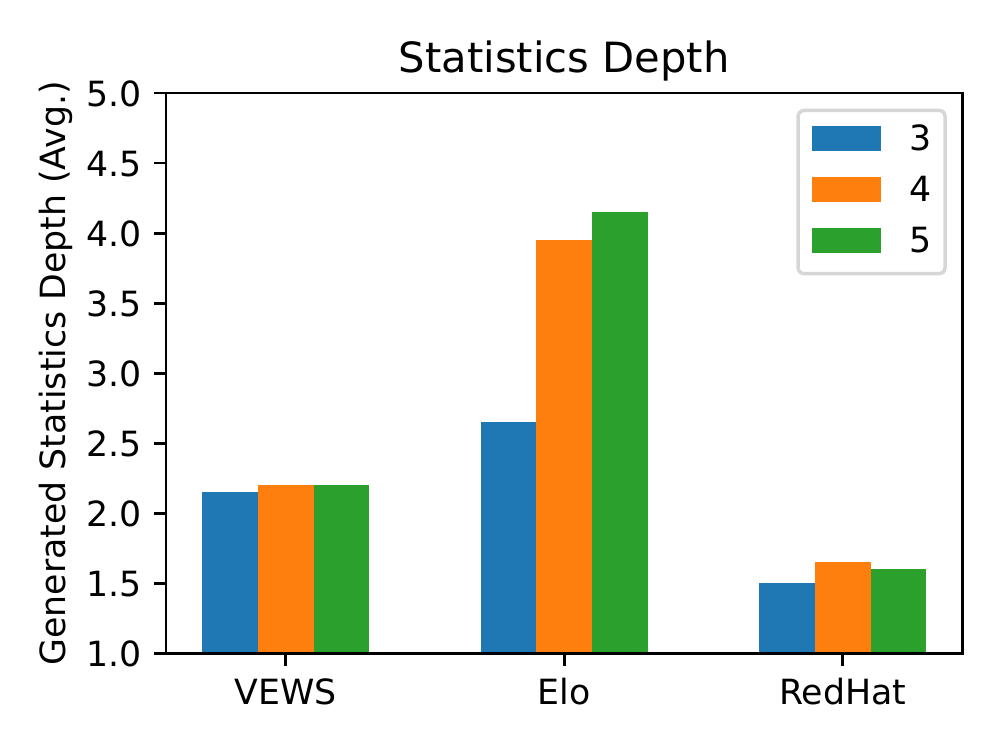}
\caption{Generated statistics depth}
\label{fig:exp_avgdepth}
\end{subfigure}
\\
\begin{subfigure}{.45\linewidth}
\centering
\includegraphics[width=\linewidth]{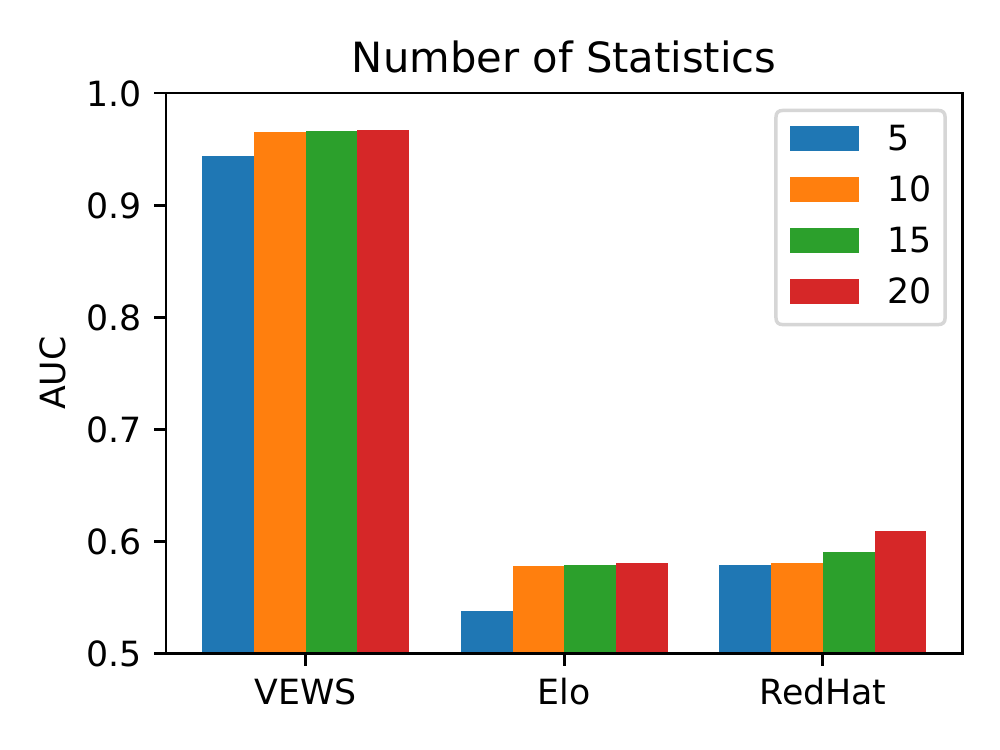}
\caption{Number of statistics $K$}
\label{fig:exp_stat}
\end{subfigure}
&
\begin{subfigure}{.45\linewidth}
\centering
\includegraphics[width=\linewidth]{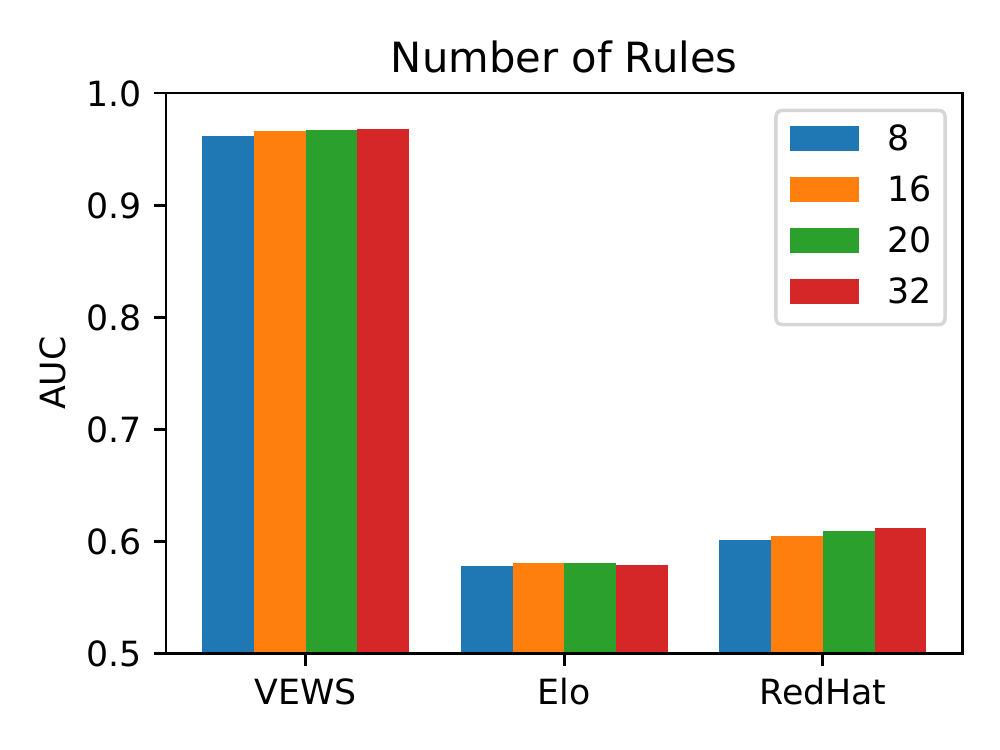}
\caption{Number of rules $R$}
\label{fig:exp_rules}
\end{subfigure}
\\
\begin{subfigure}{.45\linewidth}
\centering
\includegraphics[width=\linewidth]{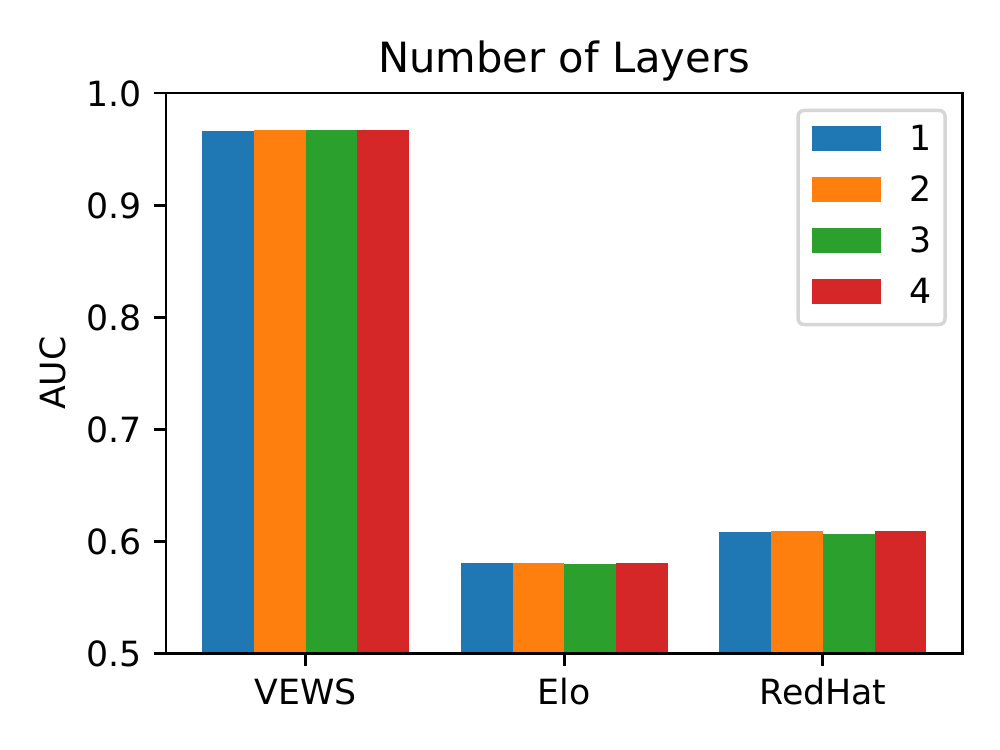}
\caption{Number of layers $L$}
\label{fig:exp_layer}
\end{subfigure} 
&
\begin{subfigure}{.45\linewidth}
\centering
\includegraphics[width=\linewidth]{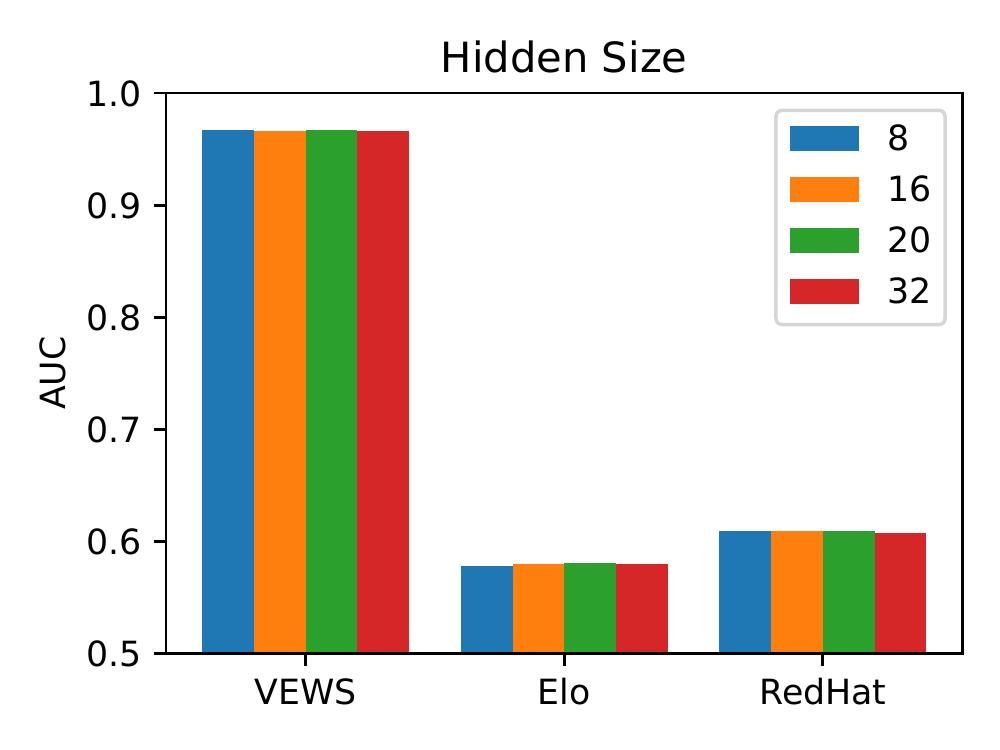}
\caption{Hidden size $H$}
\label{fig:exp_hidden}
\end{subfigure} 
\end{tabular}
\caption{Parameter study.}
\label{fig:exp_para}
\end{figure}

\begin{table*}
\caption{The learned rules and their meanings on industrial dataset.}
\label{tab:case}
\begin{tabular}{p{0.06\linewidth}|p{0.3\linewidth}|p{0.5\linewidth}}
\toprule
Weight & Rule & Meaning \\
\midrule
-0.3722 & $(\op{Sum} \circ \op{Select}_\text{cardtype=debit} > 14)$ OR $(\op{Sum} \circ \op{Select}_\text{money} > 44916)$ & Users with over 14 debit card transactions or 44916 total payment amount are less likely to be fraudulent. \\
\midrule
-0.3042 & $\op{Sum} \circ \op{Select}_\text{cardtype=credit} > 9$ & Users with over 9 credit card transactions are less likely to be fraudulent.\\
\midrule
+0.5174 & $(\op{Sum} \circ \op{Select}_\text{industry=virtual} > 9)$ AND $(\text{NOT } \op{Sum} \circ \op{Select}_\text{industry=household} > 17)$ &  Users with over 9 transactions in virtual products and no more than 17 transactions in household products are more likely to be fraudulent.\\
\midrule
+0.4564 & $(\text{NOT } \op{Sum} \circ \op{Select}_\text{money} > 4457.3)$ AND $(\text{NOT } \op{Mean} \circ \op{Select}_\text{scene=travel} > 0)$ & Users with no more than 4457.3 total payment amount and no travel-related transactions are more likely to be fraudulent.\\
\bottomrule
\end{tabular}
\end{table*}

\subsection{Experiment 2: Rule Distillation}

Now we show the results of rule distillation tasks where the goal is to distill the knowledge of teacher models into rule-based systems.

To the best of our knowledge, this paper is the first study on sequence model explanation.
There are no out-of-the-box baseline models, and most of them require fixed-length feature vectors as inputs.
In the first group of experiments, we have shown that MCTS-SG can generate informative statistics.
So we include the following alternatives with inputs being statistics generated by MCTS-SG:
(1) Lasso: We train Lasso models to mimic the outputs of teacher models and tune penalty coefficients so that the number of nonzero weights is approximately $R$.
(2) Lasso (Binary): Considering Lasso can only assign weights to the features and is not a real rule model, we use the binarized statistics (Eq. \ref{eq:binarize}) as input features.
(3) CART \cite{breiman2017classification}: This is a classic decision tree model that can learn hierarchical rules.
(4) RRL \cite{wang2021scalable}: This is the state-of-the-art neural logical model we have introduced before (Section \ref{sec:compare}).
Other logical models like SBRL \cite{yang2017scalable}, CORELS \cite{angelino2017learning}, CRS \cite{wang2020transparent} have been shown inferior to RRL and thus are not included.

We show AUC scores to test the performance of distilled rule models.
Also, we use the following score to test if  the distilled models are with fidelity to their teachers.
\begin{equation*}
\text{Fidelity} = \frac{1}{N \times (N-1)}\sum_{i\neq j} \mathbb{I}_{y_i > y_j} \odot \mathbb{I}_{\tilde{y}_i > \tilde{y}_j},
\end{equation*}
where $\odot$ is the logical XNOR operation. 
The higher the value, the more faithful the student.

We train comparing models on the training splits, and results on testing splits are summarized in Table \ref{tab:exp_main}.
We can observe that our proposed two-stage method \ours{} achieves the best performance constantly and significantly with $p$<0.05 \cite{demvsar2006statistical}.
RRL achieves comparable results, but the most complicated rule has 40 literals, which reduces its interpretability. 
On the contrary, our proposed \ours{} with two logical layers produces neat rules with at most 4 literals.

\subsection{Parameter Sensitivity}

In this section, we study the importance of several model parameters with GRU being the teacher, and show results in Figure \ref{fig:exp_para}.

First we study how the max depth of statistics $D$ affects the quality of generated statistics. We present the coefficients of multiple correlation  in Figure \ref{fig:exp_depth}, and the average depth of generated statistics in Figure \ref{fig:exp_avgdepth}.
We can observe that on Elo dataset, as we increase the max depth, the generated statistics become deeper and result in better performance.
On the contrary, on VEWS and RedHat, the generated statistics are relatively shallow even the max depth $D$ is large.
This means that our proposed statistics generation method can adaptively adjust the complexity of statistics according to datasets thanks to MCTS-based search strategy.

We also study the importance of model parameters by varying the number of statistics $K$, the number of rules $R$, the number of layers $L$, and the hidden size $H$.
From Figure \ref{fig:exp_stat} we can observe that more statistics would lead to better performance.
On VEWS and Elo datasets, it seems 20 statistics are enough, while on RedHat a better result may be achieved by generating more statistics.
The number of rules (Figure \ref{fig:exp_rules}) also positively impacts the performance.
\ours{} is insensitive to the number of layers and hidden size as shown in Figures \ref{fig:exp_layer} and \ref{fig:exp_hidden}.
These results inspire us to achieve better results with more statistics and rules (larger $K$ and $R$) and maintain the low complexity of each rule (small $L$ and $H$).

\subsection{Case Study}
\label{sec:case}

We do a case study on a sampled industrial dataset in this subsection. 
Tencent Mobile Payment Dataset is a private real-world financial fraudulent detection dataset from the Tencent Mobile Payment\footnote{The dataset is desensitized and sampled properly only for experiment purpose, and does not imply any commercial information. All personal identity information (PII) has been removed. Besides, the experiment was conducted locally on Tencent's server by formal employees.}.
The dataset includes 34.7 million transactions from 222,306 users.
Less than 4.56\% of users are labeled as fraudsters.
The teacher model is an ensemble model with manually-designed features by senior business analysts. 
A part of distilled rules is summarized in Table \ref{tab:case}.
Clearly, \ours{} successfully explains the black-box scoring model with a rule-based system, and rules are highly aligned with the industrial scenarios.

\section{Conclusion}

In this paper, we propose \ours{}, which can distill the knowledge of a user behavior sequence model into a set of rules.
An MCTS-based statistic generation method, MCTS-SG, is proposed to extract informative statistics from behavior sequences.
Then a novel neural logical model, GS-NLN, is utilized to compose rules from the statistics.
Experiments on three public datasets and one industrial dataset prove the effectiveness of our method.

As for future work, we will explore general dependency-based reward functions instead of linear correlation.
Besides, the proposed MCTS-based search method shows potential for supporting relational databases, which is also worth exploring.

\begin{acks}
This work is funded in part by the National Natural Science Foundation of China Project No. U1936213 and China Postdoctoral Science Foundation 2022M710747.
\end{acks}

\balance{
\bibliographystyle{ACM-Reference-Format}
\bibliography{ref}
}

\end{document}